\PassOptionsToPackage{numbers,sort&compress}{natbib}
\documentclass[sigconf,nonacm]{acmart}

\usepackage[utf8]{inputenc}
\usepackage[T1]{fontenc}
\usepackage{microtype}
\usepackage{booktabs}
\usepackage{tabularx}
\usepackage{array}
\usepackage{amsmath,amssymb}
\usepackage{graphicx}
\usepackage{xurl}
\usepackage{xcolor}

\setcopyright{none}
\hypersetup{
  hidelinks,
  pdftitle={Refuse, Decompose, Refresh: A Claim-Safe Protocol for Closed-Loop AI Evaluation},
  pdfauthor={Peiying Zhu; Sidi Chang}
}

\providecommand{\tightlist}{\setlength{\itemsep}{0pt}\setlength{\parskip}{0pt}}
\begin{document}
\acmshorttitle{Refuse, Decompose, Refresh}
\acmshortauthors{Zhu \& Chang}
\pagestyle{acmstyle}
\thispagestyle{firstpage}

\twocolumn[
\begin{@twocolumnfalse}
\begin{center}
{\LARGE\bfseries Refuse, Decompose, Refresh: A Claim-Safe Protocol for Closed-Loop AI Evaluation\par}
\vspace{1.0em}

\begin{tabular}{@{}c@{\hspace{2em}}c@{}}
{\large Peiying Zhu\textsuperscript{*}} & {\large Sidi Chang\textsuperscript{*\,$\dagger$}} \tabularnewline
{\small peiying@blossomai.co} & {\small schang@blossomai.co} \tabularnewline
{\small Blossom AI} & {\small Blossom AI} \tabularnewline
{\small San Francisco, CA, USA} & {\small San Francisco, CA, USA}
\end{tabular}
\vspace{1.2em}
\end{center}

\noindent\parbox{\textwidth}{
\small
\noindent\textbf{Abstract}\\[3pt]
An AI evaluation can be perfectly reproducible and still support the wrong claim. This risk is acute in closed-loop systems: the evaluated policy determines which states are visited, which components become observable, and therefore which failures can leave a measurable trace. We propose a claim-safe evaluation protocol built around three actions. \textbf{Refuse:} abstain when a clean reference stream or a matched runtime comparison lacks support. \textbf{Decompose:} report protocol execution, operational false admission, and structural hypotheses as separate decisions rather than one scientific PASS/FAIL label. \textbf{Refresh:} treat distribution-shift alarms as requests to invalidate and recompute a reference map, not as direct fault evidence.

We instantiate the protocol in an aggregate-only simulator with 24 policy components, three demand regimes, two fault-mask families, and independently seeded development and heldout data. The preregistered heldout contains 1,440 cases and 21,600 partition rows. Only 55/72 regime-component units were reference-admitted and 54/55 remained runtime-admitted, making abstention part of the result. Stable false admission was 0/20 represented components, with a one-sided exact 95\% upper bound of 0.1391 under a frozen 0.20 rule. Within admitted units, affected clean traffic predicted detection substantially better than nominal fault-cell fraction: across 540 unit-arm rows nested in 20 component clusters, the cell-minus-traffic negative-log-likelihood difference was 0.1264 nats per row, with a 95\% component-cluster interval of {[}0.0593, 0.1918{]}. A separate drift log illustrates why ``null'' must be reference-relative: clean fault-null streams triggered 15/15, 0/15, and 14/15 alarms across three regimes, while only the middle regime matched the frozen detector reference. The contribution is not a universal threshold. It is an executable contract linking observable support, statistical calibration, and the exact claim each number can justify.

}

\vspace{1.5em}
\end{@twocolumnfalse}
]

\renewcommand{\thefootnote}{\fnsymbol{footnote}}
\footnotetext[1]{Both authors contributed equally to this research.}
\footnotetext[2]{Corresponding author.}
\renewcommand{\thefootnote}{\arabic{footnote}}

\section{Introduction}\label{introduction}

Modern AI evaluation is rich in numbers and poor in claim boundaries. A benchmark score may be computed exactly, averaged correctly, and reproduced bit-for-bit while measuring a construct different from the one named in the conclusion. Measurement theory describes this as a mismatch between a theoretical construct and its operationalization \cite{manualRef01}. Benchmark critiques make the same point at scale: finite task collections are often treated as measures of general capability far beyond their validated scope \cite{manualRef02}. Multi-scenario frameworks such as HELM respond by broadening scenarios and metrics \cite{manualRef03}, while stress-test research shows that otherwise equivalent pipelines can diverge on deployment-relevant behavior \cite{manualRef04}.

Closed-loop systems add a further complication. The system being evaluated helps generate the evaluation data. A policy that never visits a state cannot reveal whether the component controlling that state is healthy. A demand change can rotate occupancy toward another region and invalidate a reference distribution without planting any fault. A large intervention can remain invisible if it affects cells the policy does not use. In such settings, forcing every scheduled case into a scalar score converts absent evidence into apparent performance.

This paper argues that a trustworthy closed-loop evaluation should be organized as a claim contract rather than a leaderboard row. The contract states:

\begin{enumerate}
\def\labelenumi{\arabic{enumi}.}
\tightlist
\item
  which evidence must exist before a score is defined;
\item
  which sampling unit supports the uncertainty statement;
\item
  which variable operationalizes exposure to the tested intervention;
\item
  which distribution defines the detector null; and
\item
  which conclusions are logically independent.
\end{enumerate}

We operationalize the contract through three verbs.

\begin{itemize}
\tightlist
\item
  \textbf{Refuse:} if the policy did not generate supported clean evidence, or if clean and current streams are not jointly supported on the same partitions, return abstention rather than a diagnostic score.
\item
  \textbf{Decompose:} keep execution integrity, operational false admission, coverage, power, and structural hypotheses in separate result fields. There is no combined scientific PASS/FAIL.
\item
  \textbf{Refresh:} when the deployed distribution departs from the detector reference, invalidate and recompute the map. An alarm is not itself a fault claim.
\end{itemize}

The empirical case study is a closed-loop target policy observed only through aggregate traces. The simulator contains 24 disjoint policy components, three separately fitted demand regimes, and two families of counterfactual component faults. A preregistered heldout tests the contract after several development analyses were frozen and downgraded where necessary. One withdrawn analysis compared exact minimum hitting set (MHS) with propagation-aware greedy selection; the methods agreed because scoped probes had already collapsed the residual choice. This negative result motivated the support-first contract, and the formal heldout deliberately tests evidence eligibility and calibration rather than MHS superiority or localization accuracy.

The results support four conclusions within the audited system. First, evaluation coverage is partial: 17 of 72 regime-component units fail the clean reference gate, and one more represented component fails the runtime gate. Second, independent false-admission calibration supports a bounded operational claim, not a zero-risk claim. Third, affected clean traffic is a better measurement of signal opportunity than nominal intervention size. Fourth, clean data from a shifted regime are fault-null but not necessarily detector-null; pooled alarm counts can therefore be badly misread.

We do not propose universal support thresholds, a general distribution-shift test, or a theorem of diagnosability. We provide an executable evaluation design and a worked audit showing why the distinction among ``not evaluated,'' ``evaluated with no signal,'' ``operationally safe,'' and ``structural hypothesis supported'' must remain visible.

\section{Evaluation claims as measurement contracts}\label{evaluation-claims-as-measurement-contracts}

\subsection{Construct, operationalization, and support}\label{construct-operationalization-and-support}

Evaluation begins with a construct: capability, reliability, safety, diagnosability, or improvement. The construct is then operationalized through tasks, observable variables, interventions, and metrics. Jacobs and Wallach emphasize that operationalization necessarily introduces assumptions and can fail construct validity even when the measurement is reliable \cite{manualRef01}. Raji et al. show how benchmark-specific measurements are routinely stretched into claims of general progress \cite{manualRef02}. HELM addresses part of this problem by reporting a portfolio of scenarios and desiderata rather than one narrow accuracy measure \cite{manualRef03}.

Our focus is a prior question: is the requested measurement defined for this case? In causal and off-policy analysis, overlap determines whether a comparison is supported \cite{manualRef05,manualRef06}. High-dimensional settings make global overlap summaries especially misleading \cite{manualRef05}. We adapt the support-first principle to aggregate system traces. The estimand is not a treatment effect or policy value; it is whether a planted component intervention could have produced a stable observable difference under the traffic generated by the policy.

Diagnosability research asks a parallel question: whether the observations available to a diagnoser can distinguish the relevant faults \cite{manualRef07}. Recent work quantifies distinguishability from data \cite{manualRef08} and makes exact versus approximate error claims explicit in probabilistic systems \cite{manualRef09}. Our protocol is narrower. It does not prove diagnosability in general; it operationalizes finite-sample support for one aggregate-trace evaluation.

\subsection{Abstention before scoring}\label{abstention-before-scoring}

Selective prediction permits a model to abstain, trading coverage against error \cite{manualRef10,manualRef11}, including under deployment shift \cite{manualRef12}. Split conformal inference illustrates the value of separating model fitting from calibration \cite{manualRef13}, while risk-controlling prediction sets attach finite-sample guarantees to a specified loss using heldout data \cite{manualRef14}. Our refusal is upstream of these methods. It does not reject a low-confidence prediction; it declares that the trace comparison lacks the observable support required to construct the prediction.

This distinction changes the denominator. A benchmark that reports success over all scheduled cases silently treats unsupported cases as failures or excludes them after seeing results. A claim-safe protocol freezes the gate, reports both conditional and operational denominators, and retains abstention as an outcome.

\subsection{Reference-relative nulls and lifecycle validity}\label{reference-relative-nulls-and-lifecycle-validity}

Dataset-shift tests are designed to make systems ``fail loudly'' when current data no longer resemble a reference distribution \cite{manualRef15}. Conformal test martingales similarly connect detected change to a retraining decision \cite{manualRef16}. Neither principle implies that an alarm on clean data is a false alarm unless the clean data are drawn from the detector\textquotesingle s null distribution.

This is easy to miss in closed-loop evaluation. ``No planted fault'' describes the simulator intervention. ``Matches the frozen reference regime'' describes the detector null. These labels can differ. A demand shift can be physically real, fault-free, and correctly detected.

\section{Case study and observable quantities}\label{case-study-and-observable-quantities}

\subsection{Closed-loop policy components}\label{closed-loop-policy-components}

The simulator evaluates a target policy under demand regimes \(\lambda_0\in\{5,7,9\}\). The policy is refitted separately in each regime. Twenty-four disjoint components are indexed by time quarter, inventory half, and market third, yielding 72 regime-component units.

For each unit, the evaluator records 15 partitions of typed aggregate traces. It does not inspect the planted fault identity when determining eligibility or stable signal. Two summaries enter the frozen signal predicate: a regional distribution distance and a mean reference-current action gap.

\subsection{Counterfactual fault families}\label{counterfactual-fault-families}

Each formal intervention moves selected, direction-changeable target-field cells by one action bucket. Outward and inward versions are evaluated separately and then combined. Selection follows one of two prespecified constructions:

\begin{itemize}
\tightlist
\item
  uniform selection draws a seeded random order over cells that can change in the requested direction;
\item
  flow-weighted selection uses seeded weighted sampling without replacement, so high-occupancy clean cells tend to appear earlier.
\end{itemize}

The families separate nominal intervention size from realized exposure. At the same selected cell fraction, flow-weighted masks can affect much more policy traffic.

For direction \(d\), define affected clean traffic

\[
\tau_d = \frac{\sum_{c\in S_d}o_c}{\sum_{c\in C}o_c},
\]

where \(o_c\) is clean occupancy, \(S_d\) is the selected fault subset, and \(C\) is the component. The bidirectional covariate is \(\tau=(\tau_{out}+\tau_{in})/2\). The matched structural covariate, \texttt{cell\_fraction}, is the arithmetic mean of the two directional selected-cell fractions.

Traffic is a candidate measurement of the causal opportunity for a fault to affect observed behavior. Cell fraction is a geometric description of how much of the component definition was edited. The formal experiment asks which better predicts stable aggregate detection.

\section{The claim-safe protocol}\label{the-claim-safe-protocol}

\subsection{Refuse: two observable support gates}\label{refuse-two-observable-support-gates}

A clean partition supports a component when its existing support count is at least 12. A regime-component unit enters the reference map when at least 14 of 15 clean reference partitions support it. Failure returns \texttt{REFERENCE\_ABSTAIN}.

Runtime eligibility is evaluated on matched pairs, not on two marginal totals. Within a partition, both the reference trace and its current counterpart must reach support 12. A unit proceeds only if this paired condition holds in at least 14 of 15 partitions; both directional evaluations must qualify before their result can be combined. Otherwise the state is \texttt{RUNTIME\_ABSTAIN}.

The joint requirement prevents a common evaluation error. Fourteen supported reference partitions and fourteen supported current partitions do not imply fourteen supported comparisons if the evidence occurs in different partitions.

\subsection{Score only eligible comparisons}\label{score-only-eligible-comparisons}

For each matched partition, signal strength is \(z=\max(\mathrm{region\_d1}/0.20, |\mathrm{mean\_action\_gap}|/0.35)\). The cutoff \(z\geq1.50\) was selected before formal execution from the fixed grid \(\{1.00,1.25,1.50,1.75,2.00\}\) and was never retuned on the formal null. A direction is stable when at least 14 of 15 matched partitions signal, and a unit-level stable detection requires both directions to be stable. The state machine is

\[
\begin{gathered}
\texttt{REFERENCE\_ABSTAIN} \rightarrow \texttt{RUNTIME\_ABSTAIN} \\
\rightarrow \texttt{SIGNAL\_ELIGIBLE} \\
\rightarrow \{\texttt{DETECTED},\texttt{NOT\_DETECTED}\}.
\end{gathered}
\]

``Not detected'' is therefore a statement about a supported comparison. It is not a synonym for ``the evaluator saw no evidence.''

Zero-occupancy faults form a separate stratum. They are not shifted by an arbitrary constant before log transformation and are not inserted into the admitted primary modeling frame.

\subsection{Calibrate false admission independently}\label{calibrate-false-admission-independently}

The null schedule uses disjoint seeds and clean-current observations. The primary safety sampling unit is the distinct physical component. The frozen statement is:

\begin{quote}
the one-sided exact 95\% upper confidence bound on component-level stable false admission is at most 0.20.
\end{quote}

The 0.20 value is an operational tolerance for this experiment, not a universal standard and not a nominal 5\% false-alarm rate. Coverage, runtime rejection, and false admission have different denominators and are reported separately.

\subsection{Decompose the decision ledger}\label{decompose-the-decision-ledger}

Table 1 defines the result grammar. The execution label can stop interpretation, but no scientific endpoint can retroactively change whether the protocol executed. Operational safety has its own label. Each structural hypothesis receives its own \texttt{PASS}, \texttt{FAIL}, or \texttt{NOT\ ESTIMABLE}. Descriptive coverage never becomes a thresholded population claim.

\begin{table*}[t]
\caption{Frozen claim ledger.}
\centering
\footnotesize
\setlength{\tabcolsep}{3pt}
\renewcommand{\arraystretch}{1.08}
\begin{tabularx}{\linewidth}{@{}>{\raggedright\arraybackslash}p{0.15\linewidth}>{\raggedright\arraybackslash}p{0.24\linewidth}>{\raggedright\arraybackslash}p{0.27\linewidth}>{\raggedright\arraybackslash}X@{}}
\toprule
Layer & Question & Allowed output & Forbidden inference \\
\midrule
execution & Did the locked protocol run and validate? & \texttt{EXECUTED} / \texttt{NOT\ EXECUTED} & execution is not scientific success \\
operational safety & Did the false-admission upper bound meet 0.20? & \texttt{SAFETY\ CONFIRMED} / \texttt{NOT\ CONFIRMED} & not a 5\% false-alarm guarantee \\
descriptive coverage & Where was evaluation supported? & counts, rates, exact intervals & no post hoc population pass threshold \\
structural hypotheses & Which operationalization predicts signal? & separate \texttt{PASS} / \texttt{FAIL} / \texttt{NOT\ ESTIMABLE} & no combined scientific label \\
maintenance & Did the detector request refresh? & alarm and refresh state & alarm is not a fault diagnosis \\
\bottomrule
\end{tabularx}
\end{table*}

This separation is a safeguard against result-driven redesign. A disappointing power result cannot justify changing the safety denominator; a high admission rate cannot compensate for false admission; a valid execution does not force structural hypotheses to pass.

\subsection{Refresh: treat shift as reference invalidation}\label{refresh-treat-shift-as-reference-invalidation}

A frozen detector monitors clean-null partitions. An alarm requests map recomputation. The formal refresh endpoint checks only that the state machine executes and reproduces the expected map hash when given the same frozen reference buffer. By construction, it cannot show that refresh repairs a genuinely stale map; it is a smoke test, not evidence of adaptation.

\section{Frozen experiment and validation}\label{frozen-experiment-and-validation}

Development data selected hypotheses and thresholds. The formal heldout uses disjoint seeds, fixed traffic targets \(\{0.15,0.30,0.50,0.75,0.95\}\), two mask families, three regimes, both directions, and 15 partitions per case. The schedule contains 1,440 cases, 21,600 partition rows, and 3,456,000 episodes. An episode is a Monte Carlo trajectory, not an independent statistical unit. Coverage begins with 72 regime-component units; safety and bootstrap inference use 20 represented physical-component clusters; and the structural fit contains 540 unit-arm rows nested within those clusters.

The primary model comparison is evaluated over 540 rows: 54 runtime-admitted units \(\times\) 5 traffic targets \(\times\) 2 families. Component-cluster bootstrap intervals use physical component as the resampling cluster. The structural endpoints are:

\begin{enumerate}
\def\labelenumi{\arabic{enumi}.}
\tightlist
\item
  \textbf{traffic over cell fraction:} lower endpoint of the two-sided 95\% interval for cell-model minus traffic-model negative log likelihood must exceed zero;
\item
  \textbf{practical family sufficiency:} the one-sided 95\% upper bound on the log-loss gain from adding family and family-by-traffic interaction must be below 0.01 nats per unit-arm row; and
\item
  \textbf{aggregate monotonicity:} total stable detections must be nondecreasing across the five traffic targets in both families.
\end{enumerate}

Independent validation recomputes admission, masks, occupancy joins, power counts, calibration bins, detector events, refresh invariants, and endpoint formulas from saved rows. It also verifies frozen hashes, schedule completeness, seed ranges, and the absence of a combined scientific label.

\subsection{Reproduction package and compute}\label{reproduction-package-and-compute}

The reproducibility artifact is available at \url{https://anonymous.4open.science/r/artifact-9f37d2/tae_2026/README.md}. It contains the frozen protocol, configuration, trajectory generator, independent validator, imported simulator source, all reference/mask/case/partition rows, and an anonymization hash map. From its unpacked root, \texttt{python3 verify.py} checks the content manifest and independently regenerates all 56 validation checks, exact bounds, model fits, 2,000 deterministic component-cluster bootstrap draws, and decision labels; it does not trust the saved endpoint summaries. Formal seeds are 200000--202399 for reference, 202400--204799 for current/fault, and 204800--204999 reserved and unused. The artifact documents every field and the exact Python environment.

Execution used one local, single-process CPU worker on an Apple M5 Pro host (18 CPU cores, 48 GB system memory); no GPU was used. The formal reference phase took 20.5 seconds and the current/fault phase 10,824.5 seconds (3.01 hours). Fifteen development executions with persisted timers total 9,099.1 seconds (2.53 hours), giving a minimum directly auditable project total of 19,944.1 seconds (5.54 CPU-hours). This excludes editing, literature work, plotting, hash-only checks, and untimed exploratory commands. Peak resident memory was not instrumented; 48 GB is the recorded host capacity. Formal saved artifacts occupy approximately 24 MB.

\section{Results}\label{results}

\subsection{Execution and refusal}\label{execution-and-refusal}

All 56 independent validation checks passed, with all 1,440 cases and 21,600 partition rows present. The execution label is \texttt{EXECUTED}.

The reference gate admitted 55/72 regime-component units (76.4\%), representing 20/24 physical components. Reference-map invalidation on independent reference evidence was 0/20, with a two-sided exact 95\% interval of {[}0.0\%, 16.8\%{]}. Runtime two-stream rejection was 1/20 represented components, or 5.0\%, with interval {[}0.1\%, 24.9\%{]}. At the unit level, 54/55 reference-admitted units remained admitted.

The 17 clean reference abstentions and one runtime abstention are evaluation results. Had the protocol forced scores on those cases, their interpretation would depend on unverified extrapolation.

\subsection{Safety is an upper bound, not a zero count}\label{safety-is-an-upper-bound-not-a-zero-count}

No represented physical component crossed the stable-signal gate in the independent null arm: 0 events over 20 components. Exact binomial inversion yields a one-sided 95\% upper limit of 0.1391, which lies below the prespecified 0.20 tolerance and therefore produces \texttt{SAFETY\ CONFIRMED}. A finer, descriptive regime-component view likewise records 0/54, with upper limit 0.0540.

The claim is bounded false admission under this sampling frame. It is not ``the evaluator has no false positives,'' and it is not a guarantee for unseen component classes or shifted nulls.

\subsection{The nominal fault-size metric measures the wrong construct}\label{the-nominal-fault-size-metric-measures-the-wrong-construct}

The traffic model outperformed the cell-fraction model by 0.1264 nats per unit-arm row in negative log likelihood. The component-cluster 95\% interval was {[}0.0593, 0.1918{]}, entirely above zero. The frozen traffic-over-cell hypothesis therefore passed.

The result has a direct measurement interpretation. Cell fraction asks how much of the component definition was changed. Affected clean traffic asks how much behavior flowed through the changed region before the fault. Only the latter measures the opportunity for the counterfactual to alter an aggregate trace.

The two mask families provide a useful stress test because they decouple these quantities. At the same traffic target, flow-weighted masks often use fewer cells. After conditioning on traffic, adding family and family-by-traffic interaction improved log loss by 0.0015 nats per unit-arm row, with a one-sided 95\% upper bound of 0.0066. This met the frozen 0.01-nat practical-sufficiency rule within these two families. However, the secondary descriptive traffic-plus-family coefficient was -0.3618 with interval {[}-0.8617, 0.0367{]}, corresponding to an odds-ratio interval {[}0.42, 1.04{]}. Because that interval extends below the frozen practical-equivalence range {[}0.5, 2.0{]}, it does not establish coefficient-scale equivalence; the \texttt{PASS} applies only to the log-loss sufficiency endpoint. The log-loss margin is only 0.0034 nats, and the percentile cluster bootstrap uses 20 represented components; undercoverage could favor this conclusion.

Aggregate bidirectional detection counts were nondecreasing across traffic targets:

\begin{itemize}
\tightlist
\item
  uniform: 0, 17, 37, 52, 54;
\item
  flow-weighted: 0, 13, 35, 49, 54.
\end{itemize}

This is not a universal causal law. It is a frozen structural result within admitted units, three regimes, two mask families, and the tested fault construction.

\begin{table*}[t]
\caption{Formal results retain separate meanings.}
\centering
\small
\begin{tabularx}{\textwidth}{@{}>{\raggedright\arraybackslash}p{0.19\textwidth}>{\raggedright\arraybackslash}p{0.27\textwidth}>{\raggedright\arraybackslash}X@{}}
\toprule
Item & Result & Claim supported \\
\midrule

protocol execution & 56/56 validation checks & frozen run is interpretable \\
reference admission & 55/72 units; 20/24 components & descriptive evaluation coverage \\
runtime rejection & 1/20 components & one supported component became incomparable \\
stable false admission & 0/20; upper 0.1391 & frozen operational safety criterion met \\
traffic vs cell fraction & 0.1264 {[}0.0593, 0.1918{]} nats/row & traffic is better within admitted units \\
family gain beyond traffic & 0.0015; upper 0.0066 & practical sufficiency within two families \\
traffic-target counts & nondecreasing in both families & aggregate monotonicity hypothesis passed \\
\bottomrule
\end{tabularx}
\end{table*}

\subsection{A clean stream is not automatically a detector null}\label{a-clean-stream-is-not-automatically-a-detector-null}

The drift log scored 45 clean-current partitions, 15 per demand regime. Pooled, the detector alarmed on 29/45 partitions. Read without the reference definition, this looks like a 64\% false-alarm rate. It is not.

The detector\textquotesingle s frozen reference statistics came from development data with \(\lambda_0=7\). Stratifying the 45 fault-null partitions gives Table 3.

\begin{table*}[t]
\caption{Detector alarms are reference-relative.}
\centering
\small
\begin{tabularx}{\textwidth}{@{}>{\raggedright\arraybackslash}p{0.18\textwidth}>{\raggedright\arraybackslash}Xr@{}}
\toprule
Current regime & Relation to frozen detector reference & Alarms \\
\midrule

\(\lambda_0=5\) & demand-shifted & 15/15 \\
\(\lambda_0=7\) & reference-matched & 0/15 \\
\(\lambda_0=9\) & demand-shifted & 14/15 \\
\bottomrule
\end{tabularx}
\end{table*}

The middle row is the clean reference-matched null. The outer rows are physically fault-free but distributionally shifted. Their alarms are consistent with the detector\textquotesingle s intended role: request map refresh when the current regime no longer resembles the frozen reference. The experiment did not preregister a detector-sensitivity threshold, so this stratified pattern is descriptive rather than a confirmatory power claim.

The preregistered false-admission endpoint in Section 6.2 is the relevant operational false-positive measurement. Detector alarms and stable fault admissions answer different questions.

\section{What this changes about AI evaluation practice}\label{what-this-changes-about-ai-evaluation-practice}

\subsection{Publish the denominator-generating mechanism}\label{publish-the-denominator-generating-mechanism}

Evaluation papers routinely report the denominator after filtering without explaining how the system\textquotesingle s own behavior caused inclusion. In closed-loop systems, admission is endogenous to the policy. A trustworthy report should publish scheduled, reference-admitted, runtime-admitted, and scored denominators, plus the rule connecting them.

\subsection{Validate the exposure variable, not only the score}\label{validate-the-exposure-variable-not-only-the-score}

Metrics often inherit convenient proxies: number of edited tokens, tool calls, cells, steps, or changed parameters. These are structural sizes, not necessarily causal exposures. Competing operationalizations should be compared on heldout predictive or decision relevance. Here, the nominal fault-size metric loses decisively to traffic despite appearing natural.

\subsection{Name the null distribution}\label{name-the-null-distribution}

``Clean,'' ``benign,'' ``no fault,'' and ``in distribution'' are not synonyms. Every alarm rate should state the reference distribution and the level at which exchangeability is assumed. Pooling distinct regimes can convert correct shift detection into an apparent false-positive crisis.

\subsection{Make refusal visible}\label{make-refusal-visible}

Abstention should be reported alongside accuracy or detection, not buried in preprocessing. Conditional performance answers ``how well did the evaluator work where it declared evidence sufficient?'' Operational performance answers ``how often did the full system return a correct usable result among scheduled opportunities?'' Both are needed.

\subsection{Replace global verdicts with a claim ledger}\label{replace-global-verdicts-with-a-claim-ledger}

A single PASS/FAIL creates logical coupling among unrelated endpoints. The alternative is small and mechanical: one execution state, one operational-safety decision, descriptive coverage, and one label per structural hypothesis. Each headline claim should be paired with its nearest unsupported case.

\section{Broader impacts}\label{broader-impacts}

The intended benefit is safer evaluation practice. Making support, denominators, reference distributions, and refusal visible can prevent an exact score from being used to justify an unsupported diagnosis. The protocol may be useful in monitoring, auditing, or safety testing closed-loop systems, especially when operators might otherwise mistake distribution shift for a fault or silently score unobservable cases.

The same machinery can cause harm if its conditional claims are promoted into deployment guarantees. A frozen threshold may be copied into a new domain without validation; a passed safety gate may create false assurance; and abstention may concentrate on rare, low-traffic components, leaving failures that affect low-volume users or conditions systematically unevaluated. Frequent shift alarms can also impose operational cost or be misused to justify unnecessary intervention. In a real system, releasing detailed occupancy maps or traces could expose sensitive behavior even when the evaluator is statistically sound.

Mitigations follow from the claim ledger: report coverage and abstention by deployment-relevant strata, audit who or what is missing, keep alarms separate from fault evidence, require domain-specific recalibration and human review before action, and apply access controls or aggregation to real logs. This release contains only simulator-generated traces, no human-subject or personal data, and no deployable model. Those facts reduce privacy and direct misuse risk but do not establish fairness or safety for any real deployment.

\section{Limitations}\label{limitations}

The thresholds were developed for one simulator and should not be transferred mechanically. The study uses 24 components, three demand regimes, two mask families, aggregate traces, and fixed support counts. Systems without clean reference streams need another design.

The formal heldout does not establish localization accuracy. It evaluates whether aggregate comparisons are supported and whether planted interventions yield stable signal as traffic increases. It also does not validate the drift detector as a sensitivity-optimized change detector.

The family-sufficiency result rests on approximately 20 component clusters and a percentile cluster bootstrap. Its 0.0034-nat margin may be vulnerable to finite-cluster undercoverage. We preserve family-specific calibration rows so readers can judge the aggregation.

Refresh was exercised with the original locked buffer, so the map could not change by design. This validates control flow and the map-hash invariant; it provides no empirical evidence that fresh post-shift data would repair an obsolete map.

Finally, this case study demonstrates a claim-safe architecture, not universal construct validity. External validity still requires new systems, regimes, fault mechanisms, and deployment conditions.

\section{Conclusion}\label{conclusion}

Trust in AI evaluation does not follow from precise arithmetic alone. The evaluator must show that the case was observable, that the compared streams were supported, that its metric operationalizes the intended mechanism, and that its null matches the reported alarm interpretation.

Our closed-loop case study makes these requirements executable. The protocol refuses unsupported cases, independently calibrates stable false admission, compares alternative exposure measurements, distinguishes fault-null from detector-null streams, and reports execution, safety, and structural hypotheses separately. The formal results show both the value and the cost of this discipline: coverage is incomplete, uncertainty remains visible, and one apparently alarming pooled count becomes interpretable only after restoring the reference definition.

A trustworthy evaluator should sometimes refuse to score, should never compress independent claims into one verdict, and should refresh its reference when the world moves. Those behaviors are not signs of evaluation weakness. They are evidence that the evaluation knows what its numbers mean.

\bibliographystyle{plain}
\bibliography{references}

\end{document}